\documentclass[conference]{IEEEtran}
\usepackage{cite}
\usepackage{times}
\usepackage{graphicx}
\usepackage{latexsym}

\usepackage{url}
\usepackage{amsmath}
\usepackage{algorithm}
\usepackage{algorithmic}
\usepackage{caption}
\usepackage{float}
\usepackage{array}
\usepackage{multirow}
\usepackage{graphicx}
\usepackage{verbatim}
\usepackage{hhline}
\usepackage{color}

\newcommand{\etal}{\textit{et al}}

\usepackage{amsfonts}
\makeatletter
\def\amsbb{\use@mathgroup \M@U \symAMSb}
\makeatother

\DeclareSymbolFont{bbold}{U}{bbold}{m}{n}
\DeclareSymbolFontAlphabet{\mathbbold}{bbold}
\DeclareMathOperator*{\argmax}{arg\,max}
\DeclareMathOperator*{\argmin}{arg\,min}

\newcommand{\specialcell}[2][c]{%
\begin{tabular}[#1]{@{}c@{}}#2\end{tabular}}

\ifCLASSINFOpdf
\else
\fi
\begin{document}
\title{Constraint-Based Clustering Selection}

\author{\IEEEauthorblockN{Toon Van Craenendonck}
\IEEEauthorblockA{Department of Computer Science\\
KU Leuven, Belgium\\
 toon.vancraenendonck@cs.kuleuven.be}
\and
\IEEEauthorblockN{Hendrik Blockeel}
\IEEEauthorblockA{Department of Computer Science \\ KU Leuven, Belgium \\
hendrik.blockeel@cs.kuleuven.be}
}


%Telephone: (800) 555--1212, Fax: (888) 555--1212}


\maketitle

\begin{abstract}
Semi-supervised clustering methods incorporate a limited amount of supervision into the clustering process. Typically, this supervision is provided by the user in the form of pairwise constraints. Existing methods use such constraints in one of the following ways: they adapt their clustering procedure, their similarity metric, or both. All of these approaches operate within the scope of individual clustering algorithms. In contrast, we propose to use constraints to choose between clusterings generated by very different unsupervised clustering algorithms, run with different parameter settings. We empirically show that this simple approach often outperforms existing semi-supervised clustering methods.   
\end{abstract}


% \fi
\IEEEpeerreviewmaketitle

\section{Introduction}
%The traditional page limit for ECAI long papers is {\bf 6 (six)} pages
%in the required format. The traditional page limit for short
%submissions is {\bf 2} pages.
%
%However, these page limits may change from one ECAI to
%another. Consult the most recent Call For Papers (CFP) for the most
%up-to-date page limits.

Clustering is one of the core tasks in data analysis \cite{Jain2010}. It is inherently subjective, as users may prefer very different clusterings of the same data \cite{Caruana06metaclustering,ScienceOrArt}. Semi-supervised clustering \cite{Wagstaff01constrainedk-means,xing2002distance} aims to deal with this subjectivity by allowing the user to specify background knowledge, often in the form of pairwise constraints that indicate whether two instances should be in the same cluster or not.

Traditional approaches to semi-supervised (or constraint-based) clustering use constraints in one of the following three ways. First, one can modify an existing clustering algorithm to take them into account. This approach is taken in COP-KMeans \cite{Wagstaff01constrainedk-means}, one of the first clustering algorithms able to deal with pairwise constraints. Second, one can learn a distance metric based on the constraints \cite{xing2002distance}, after which the metric is used within a traditional clustering algorithm. Third, one can combine the above two approaches and develop so-called hybrid methods \cite{Bilenko2004}. 

Our approach to constraint-based clustering is quite different from existing methods, and does not fit in any of these three categories. It is motivated by the well-known fact that different algorithms may produce very different clusterings of the same data \cite{Estivill-Castro2002}, and even within one algorithm, different parameter settings may yield different clusterings. This implies that selecting a clustering algorithm and tuning its parameter settings is crucial to obtain a good clustering.

 We propose to use constraints to solve these tasks: to find an appropriate clustering, we first generate a set of clusterings using several unsupervised algorithms, with different hyperparameter settings, and afterwards select from this set the clustering that satisfies the largest number of constraints. Our experiments show that, surprisingly, this simple constraint-based selection approach often yields better clusterings than existing semi-supervised algorithms. This shows that it is more important to use an algorithm of which the inherent bias matches a particular problem, than to modify the optimization criterion of any individual algorithm to take the constraints into account. We also present a method for selecting the most informative constraints first, which further increases the usefulness of our approach.
 
The remainder of this paper is structured as follows. In Section \ref{sec:background} we give some background on semi-supervised clustering, and algorithm and hyperparameter selection for clustering. Section \ref{sec:COBS_g} presents our approach to using pairwise constraints in clustering, which we call COBS (for Constraint-Based Selection). In Section 4 we describe how COBS can be extended to actively select informative constraints. We conclude in Section \ref{sec:conclusion}.

\section{Background}
\label{sec:background}

We first describe related work on semi-supervised clustering. As our approach consists of using constraints to choose an algorithm and tune its parameters, we also discuss related work on meta-learning for clustering, as well as on algorithm and hyperparameter selection. \\

\textbf{Semi-supervised clustering algorithms} allow the user to incorporate a limited amount of supervision into the clustering procedure. Several kinds of supervision have been proposed, one of the most popular ones being pairwise constraints. Must-link (ML) constraints indicate that two instances should be in the same cluster, cannot-link (CL) constraints that they should be in different clusters. Most existing semi-supervised approaches use such constraints within the scope of an individual clustering algorithm. COP-KMeans \cite{Wagstaff01constrainedk-means}, for example, modifies the clustering assignment step of K-means: instances are assigned to the closest cluster for which the assignment does not violate any constraints. Similarly, the clustering procedures of DBSCAN \cite{Ruiz2007,SSDBSCAN,Campello}, EM \cite{Shental2004} and spectral clustering \cite{Rangapuram2012,Wang2014} have been extended to incorporate pairwise constraints. Another approach to semi-supervised clustering is to learn a distance metric based on the constraints \cite{xing2002distance,BarHillel2003LearningDF,Davis:2007:IML:1273496.1273523}. Xing \etal .\ \cite{xing2002distance}, for example, propose to learn a Mahalanobis distance by solving a convex optimization problem in which the distance between instances with a must-link constraint between them is minimized, while simultaneously separating instances connected by a cannot-link constraint. Hybrid algorithms, such as MPCKMeans \cite{Bilenko2004}, combine metric learning with an adapted clustering procedure. \\ 

\textbf{Meta-learning and algorithm selection} have been studied extensively in supervised learning \cite{Brazdil,AutoWEKA}, but much less clustering. There is some work on building meta-learning systems that recommend clustering algorithms \cite{DeSouto2008,Ferrari2015}. However, these systems do not take hyperparameter selection into account, or any form of supervision. More related to ours is the work of Caruana \etal .\ \cite{Caruana06metaclustering}. They generate a large number of clusterings using K-means and spectral clustering, and cluster these clusterings. This meta-clustering is presented to the user as a dendrogram. Here, we also generate a set of clusterings, but afterwards we select from that set the most suitable clustering based on pairwise constraints. The only other work, to our knowledge, that has explored the use of pairwise constraints for algorithm selection is that by Adam and Blockeel \cite{Adam}. They define a meta-feature based on constraints, and use this feature to predict whether EM or spectral clustering will perform better for a dataset. While their meta-feature attempts to capture one specific property of the desired clusters, i.e.\ whether they overlap, our approach is more general and allows selection between any clustering algorithms. \\

Whereas algorithm selection has received little attention in clustering, several methods have been proposed for \mbox{\textbf{hyperparameter selection}}.

One strategy is to run the algorithm with several parameter settings, and select the clustering that scores highest on an internal quality measure \cite{Arbelaitz2013,SAM:SAM10080}. Such measures try to capture the idea of a ``good'' clustering. A first limitation is that they are not able to deal with the inherent subjectivity of clustering, as they do not take any external information into account. Furthermore, internal measures are only applicable within the scope of individual clustering algorithms, as each of them comes with its own bias \cite{ScienceOrArt}. For example, the vast majority of them has a preference for spherical clusters, making them suitable for K-means, but not for e.g.\ spectral clustering and DBSCAN. 

Another strategy for parameter selection in clustering is based on stability analysis \cite{BenHur,vonLuxburgStability,Ben-David2006}. A parameter setting is considered to be stable if similar clusterings are produced with that setting when it is applied to several datasets from the same underlying model. These datasets can for example be obtained by taking subsamples of the original dataset \cite{BenHur,Lange}. In contrast to internal quality measures, stability analysis does not require an explicit definition of what it means for a clustering to be good. Most studies on stability focus on selecting parameter settings in the scope of individual algorithms (in particular, often the number of clusters). 

Additionally, one can also avoid the need for explicit parameter selection. In self-tuning spectral clustering \cite{Zelnik-manor04self-tuningspectral}, for example, the affinity matrix is constructed based on local statistics and the number of clusters is estimated using the structure of the eigenvectors of the Laplacian.

A key distinction with COBS is that none of the above methods takes the subjective preferences of the user into account. We will compare our constraint-based selection strategy to some of them in the next section.
 
\section{Constraint-based clustering selection}
\label{sec:COBS_g}

Algorithm and hyperparameter selection are difficult tasks in an entirely unsupervised setting, mainly due to the lack of a well-defined way to estimate the quality of clustering results \cite{Estivill-Castro2002}. We propose to use constraints for this purpose, and estimate the quality of a clustering as the number of constraints that it satisfies. This quality estimate allows us to do a search over unsupervised algorithms and their parameter settings, as described in Algorithm \ref{COBS_random_algo}. We use a basic grid search, but in principle also more advanced optimization strategies could be used \cite{AutoWEKA,Hutter2011}. We assume that we are given a set of must-link constraints \emph{ML}, where $(i,j) \in$ \emph{ML} indicates that instances $x_i$ and $x_j$ should be in the same cluster. Similarly, we are given a set of cannot-link constraints \emph{CL}, where $(i,j) \in$ \emph{CL} indicates that $x_i$ and $x_j$ should be in different clusters. A clustering maps instances (through their index) to their cluster label, i.e.\ $c[i] = k$ indicates that in clustering $c$, $x_i$ is an element of cluster $k$. The indicator function $\amsbb{I}$ has value one if the enclosed expression is true, zero otherwise. We select the ``best'' solution from a set of clusterings as the one satisfying the largest number of constraints (in case of a tie, we select randomly from the involved clusterings). 

\begin{comment}
A user who wants to cluster a dataset has to choose a clustering algorithm, and usually also set some hyperparameters.
Several strategies can be used to solve the equivalent problem in supervised learning \cite{AutoWEKA}, which all rely on the availability of a performance criterion to optimize. An example is basic grid search: different algorithms are run with several values for their hyperparameters, and afterwards the best model is selected as the one that scores highest on the performance criterion.\\
Such search strategies cannot be used in unsupervised clustering, as we do not have a well-defined performance criterion to optimize. In particular, without any supervision we do not have a good way to compare clusterings that are generated by different algorithms.\\
We argue that if we are given pairwise constraints, it becomes possible to estimate clustering performance: we can count the number of constraints that a clustering satisfies. This allows us to do a grid search over unsupervised clustering algorithms and their hyperparameters, as described in Algorithm \ref{COBS_random_algo}.
\end{comment}

\thickmuskip=0mu

\renewcommand{\thealgorithm}{1}

\begin{algorithm}
\caption{Constraint-based selection (COBS)}
\label{COBS_random_algo}
\begin{algorithmic}[1]
\REQUIRE $D$: a dataset\\ \ \ \ \ \ \  \emph{ML}: set of must-link constraints\\ \ \ \ \ \ \ \emph{CL}: set of cannot-link constraints 
\ENSURE a clustering of $D$
\STATE Generate a set of clusterings $C$ by varying the hyperparameters of several unsupervised clustering algorithms
\RETURN  $\displaystyle \argmax_{c \in C}\big(\!\!\!\!\sum_{(i,j) \in \text{\emph{ML}}}\!\!\!\!\amsbb{I}[c[i]=c[j]]+\!\!\!\!\sum_{(i,j)\in\text{\emph{CL}}}\!\!\!\!\amsbb{I}[c[i]\neq c[j]]\big)$ 

\end{algorithmic}    
\end{algorithm}

COBS is motivated by the following two observations.

First, it is commonly accepted that no single algorithm performs best on all clustering problems: each algorithm comes with its own bias, which may match a particular problem to a greater or lesser degree \cite{Estivill-Castro2002}. Traditional semi-supervised approaches use constraints within the scope of an individual algorithm. By doing so, they can change the bias of the algorithm, but only to a certain extent. For instance, using constraints to learn a Mahalanobis distance allows K-means to find ellipsoidal clusters, rather than spherical ones, but still does not make it possible to find non-convex clusters. In contrast, by using constraints to choose between clusterings generated by very different algorithms, COBS aims to select the most suitable one from a diverse range of biases. 

Second, it is also widely known that within a single clustering algorithm the choice of the hyperparameters can strongly influence the clustering result. Consequently, choosing a good parameter setting is crucial. Currently, a user can either do this manually, or use one of the selection strategies discussed in section \ref{sec:background}. Both options come with significant drawbacks. Doing parameter tuning manually is time-consuming, given the often large number of combinations one might try. Existing automated selection strategies avoid this manual labor, but can easily fail to select a good setting as they do not take the user's preferences into account. For COBS, parameters are an asset rather than a burden. They allow generating a large and diverse set of clusterings, from which we can select the most suitable solution with a limited number of pairwise constraints. \\

Although our approach is very simple, it does not appear to have been studied before, neither as a way to incorporate constraints into clustering, nor as a way to select clustering algorithms and their parameter settings (despite the substantial body of research on both constraint-based clustering and hyperparameter selection). 

\subsection*{Research questions}
In the remainder of this section, we aim to answer the following questions:
\begin{itemize}
	\item[\textbf{Q1}] How does COBS, for \emph{hyperparameter selection only}, compare to unsupervised hyperparameter selection methods?
	\item[\textbf{Q2}] How does COBS, for \emph{hyperparameter selection only}, compare to existing semi-supervised clustering algorithms?
	\item[\textbf{Q3}] How does COBS, for \emph{both algorithm and hyperparameter selection}, compare to existing semi-supervised algorithms?
	\item[\textbf{Q4}] Can we improve COBS by using semi-supervised algorithms to generate clusterings, instead of unsupervised ones? 
\end{itemize}

Although our selection strategy is also related to meta-clustering \cite{Caruana06metaclustering}, an experimental comparison would be difficult as meta-clustering produces a dendrogram of clusterings for the user to explore. The user can traverse this dendrogram to obtain a single clustering, but the outcome of this process is highly subjective. COBS works with pairwise constraints, therefore we compare to other methods that do the same.

\begin{comment}
In the remainder of this section, we aim to answer the following questions:
\begin{enumerate}
	\item[\textbf{Q1}] How does COBS for hyperparameter selection compare to unsupervised hyperparameter selection methods?
	\item[\textbf{Q2}] How does COBS for hyperparameter selection compare to existing semi-supervised clustering algorithms?
	\item[\textbf{Q3}] How does COBS, for both algorithm and hyperparameter selection, compare to existing semi-supervised algorithms?
	\item[\textbf{Q4}] Can we improve COBS by using semi-supervised algorithms to generate clusterings, instead of unsupervised ones?	 
\end{enumerate}
Q1 and Q2 consider the use of constraints for hyperparameter selection only, whereas Q3 and Q4 consider both algorithm and hyperparameter selection.
\end{comment}

\subsection*{Experimental methodology}
To answer our research questions we perform experiments with 10 UCI classification datasets, listed in Table \ref{table:datasets}. These have also been used in several other studies on semi-supervised clustering  \cite{Bilenko2004,Xiong2014}. The optdigits389 dataset is a subset of the UCI handwritten digits dataset, containing only digits 3, 8 and 9  \cite{Bilenko2004,Mallapragada2008}.  The classes are assumed to represent the clusters of interest. We evaluate how well the returned clusters coincide with them by computing the Adjusted Rand Index (ARI) \cite{ARI}, which is a commonly used measure for this; 0 means that the clustering is not better than random, 1 is a perfect match.  In our experiments with semi-supervised clustering, we always repeat the following steps 25 times and report average results:

\begin{enumerate}
	\item Randomly partition the full dataset into 70\% (``potential supervision set'') and 30\% (``left-out set'').
	\item Generate $c$ pairwise constraints ($c$ is a parameter) by repeatedly selecting two random instances from the supervision set, and adding a must-link constraint if they belong to the same class, and a cannot-link otherwise. 
	\item Apply COBS to the full dataset to obtain a clustering.
	\item Evaluate the clustering by calculating the ARI on all objects that were not involved in any constraints.
\end{enumerate}
We avoid including pairs in the evaluation that were among the given constraints, as this would be the equivalent of testing on the training set. \\

\begin{table}
  \footnotesize
   \caption{{\footnotesize Datasets used in the experiments. Duplicate instances and instances with missing values are removed.  }}\label{table:datasets}
	\centering  
  \begin{tabular}{|c|c|c|c|}
 \hline
     \textbf{dataset} & \textbf{\# instances} & \textbf{\# features} & \textbf{\# classes} \\
    \hline
    wine  & 178 & 13 & 3 \\
	dermatology  & 358 & 33 & 6 \\
	iris  & 147 & 4 & 3 \\
		ionosphere  &  350 & 34 & 2 \\
			breast-cancer-wisconsin  &  449 & 32 & 2 \\
	ecoli &  336 & 7 & 8 \\
	optdigits389   &  1151 & 64 & 3 \\
	segmentation  &  2100 & 19 & 7 \\
	glass  &  214 & 10 & 7 \\
	hepatitis  &  112 & 19 & 2 \\
	\hline
	\end{tabular}
       % Add 'table' caption
\end{table}

We use K-means, DBSCAN and spectral clustering to generate clusterings in step one of Algorithm \ref{COBS_random_algo}, as they are common representatives of different types of algorithms (we use implementations from scikit-learn \cite{PedregosaF.andVaroquauxG.andGramfortA.andMichel2011}). The hyperparameters are varied in the ranges specified in Table \ref{table:algos}. In particular, for each dataset we generate 180 clusterings using K-means (for each number of clusters we store the clusterings obtained with 20 random initializations), 351 using spectral clustering and 400 using DBSCAN, yielding a total of 931 clusterings. For discrete parameters, clusterings are generated for the complete range. For continuous parameters, clusterings are generated using 20 evenly spaced values in the specified intervals. For the $\epsilon$ parameter used in DBSCAN, the lower and upper bounds are the minimum and maximum pairwise distances between instances (referred to as $\min (d)$ and $\max (d)$ in Table \ref{table:algos}).

All datasets are normalized by rescaling each feature to the range $[0,1]$. We use the Euclidean distance for all unsupervised algorithms.

\begin{table}
\label{}
    \captionsetup{justification=centering}
  \begin{center}
  \footnotesize
      \captionof{table}{{\footnotesize Algorithms used, the hyperparameters that were varied, their corresponding ranges and the hyperparameter selection methods used in Q1 }}\label{table:algos}% Add 'table' caption
  \begin{tabular}{|c|c|c|c|}
 \hline
     \textbf{Algorithm} & \textbf{Param.} & \textbf{Range} & \textbf{Selection method} \\
    \hline
	k-means  & \specialcell{$K$} & $[2,10]$ & silhouette index\\
	\hline
	DBSCAN  &  \specialcell{$\epsilon$\\ $minPts$} & \specialcell{$[\min (d), \max (d)]$ \\ $[2,20]$}  & DBCV index\\
	\hline
	spectral& \specialcell{$K$ \\ $k$ \\  $\sigma$}  & \specialcell{$[2,10]$ \\ $[2,20]$ \\ $[0.01,5.0]$}  & \specialcell{self-tuning \\ spectral clustering } \\
	\hline
	\end{tabular}
    
  \end{center}
\end{table}

\subsection*{Q1: COBS vs. unsupervised hyperparameter tuning}
\label{sec:selecting_hyperparameters}
To evaluate hyperparameter selection for individual algorithms, we use Algorithm \ref{COBS_random_algo} with $C$ a set of clusterings generated using one particular algorithm (K-means, DBSCAN or spectral). We compare COBS to state of the art unsupervised selection strategies. As there is no single method that can be used for all three algorithms, we use three different approaches, which are briefly described next.\\

\textbf{K-means} has one hyperparameter: the number of clusters $K$. A popular method to select $K$ in K-means is by using internal clustering quality measures \cite{SAM:SAM10080,Arbelaitz2013}. K-means is ran for different values of K (and in this case also for different random seeds), and afterwards the clustering that scores highest on such an internal measure is chosen. In our setup, we generate 20 clusterings for each $K$ by using different random seeds. We select the clustering that scores highest on the silhouette index \cite{Rousseeuw1987}, which was identified as one of the best internal criteria by Arbelaitz \etal .\ \cite{Arbelaitz2013}. 

\textbf{DBSCAN} has two parameters: $\epsilon$, which specifies how close points should be to be in the same neighborhood, and $minPts$, which specifies the number of points that are required in the neighborhood to be a core point. Most internal criteria are not suited for DBSCAN, as they assume spherical clusters, and one of the key characteristics of DBSCAN is that it can find clusters with arbitrary shape. One exception is the Density-Based Cluster Validation (DBCV) score \cite{Moulavi2014}, which we use in our experiments.

\textbf{Spectral clustering} requires the construction of a similarity graph, which can be done in several ways \cite{vonLuxburg}. If a $k$-nearest neighbor graph is used, $k$ has to be set. For graphs based on a Gaussian similarity function, $\sigma$ has to be set to specify the width of the neighborhoods. Also the number of clusters $K$ should be specified. Self-tuning spectral clustering \cite{Zelnik-manor04self-tuningspectral} avoids having to specify any of these parameters, by relying on local statistics to compute different $\sigma$ values for each instance, and by exploiting structure in the eigenvectors to determine the number of clusters. This approach is different from the one used for K-means and DBSCAN, as here we do not generate a set of clusterings first, but instead hyperparameters are estimated directly from the data.\\

 \begin{table*}[t]
\label{}
    \captionsetup{justification=centering}

  \begin{center}
  \setlength\tabcolsep{1.5pt} %
  	\caption{ We first show the ARIs obtained with unsupervised vs. constraint-based hyperparameter selection (columns marked Q1). Next, we show the ARIs obtained with the semi-supervised variants, with several hyperparameter selection methods (columns marked Q2). For semi-supervised results 50 constraints were used, and the average of 25 runs is shown. SI refers to the silhouette index, STS to self-tuning spectral clustering, FOSC to FOSC-OpticsDend and eigen to the eigengap method. }\label{table:hyperparam_selection} 
  	  {\footnotesize

    \begin{tabular}{|c|}
 \hline
     \multirow{3}{*}{\textbf{dataset}}   \\ 
     \\ 
     \\ \hline
     wine   \\
	dermatology    \\
	iris   \\
	ionosphere  \\
	breast-cancer-wisconsin    \\
	ecoli   \\
	optdigits389 \\
	segmentation \\
	hepatitis    \\
	glass   \\
	
	\hline
	\end{tabular}   \setlength\tabcolsep{4pt} %
  \begin{tabular}{|c|c|c|c|c|}
 \hline
     \multicolumn{2}{|c|}{\textbf{K-means}} & \multicolumn{3}{c|}{\textbf{MPCKMeans}} \\ 
      \multicolumn{2}{|c|}{\emph{Q1}} & \multicolumn{3}{c|}{\emph{Q2}} \\ \hline 
     SI & COBS & SI & NumSat & CVCP  \\     \hline
 \underline{0.85} & 0.81 & \textbf{0.86} & 0.68 & 0.70  \\
	0.57 & \underline{\textbf{0.84}} & 0.59 & 0.46 & 0.42  \\
	 0.56 & \underline{0.66} & 0.62 & \textbf{0.72} & 0.65   \\
	 \underline{\textbf{0.27}} & 0.24 & 0.24 & 0.19 & 0.17 \\
	 \underline{\textbf{0.73}} & 0.67 & \textbf{0.73} & \textbf{0.73} & 0.71 \\
	0.04 & \underline{0.62} & \textbf{0.70} & 0.51 & 0.45 \\
	0.49 & \textbf{\underline{0.79}} & 0.58 & 0.28 & 0.49\\
	0.10 & \textbf{\underline{0.51}} & 0.38 & 0.19 & 0.28 \\
	 \underline{0.19} & 0.18 & \textbf{0.25} & 0.18 & 0.14  \\
	\underline{0.23} & 0.2 & \textbf{0.24} & 0.17 & 0.20   \\
		\hline
	\end{tabular}
  \begin{tabular}{|c|c|c|}
 \hline
     \multicolumn{2}{|c|}{\textbf{DBSCAN}} & \textbf{FOSC}  \\ 
          \multicolumn{2}{|c|}{\emph{Q1}} & \emph{Q2}  \\ \cline{1-2}
     DBCV & COBS &   \\ 
    \hline
 0.32 & \underline{0.36} & \textbf{0.53} \\
	    0.37 & \underline{0.40} & \textbf{0.76} \\	  
	  \underline{0.56} & 0.50 & \textbf{0.80} \\
	      0.05 & \textbf{\underline{0.66}} & -0.04 \\
	       0.65 & \underline{\textbf{0.72}} & 0.53 \\
	  0.03 & \underline{0.44} &  \textbf{0.56}    \\
	 0.00 & \underline{0.27} & \textbf{0.55}  \\
	  0.24 & \underline{0.37} & \textbf{0.54}  \\
	  -0.13 & \underline{0.02} & \textbf{0.23} \\
	  0.01 & \underline{0.14} & \textbf{0.15} \\
	 	\hline
	\end{tabular}
  \begin{tabular}{|c|c|c|c|c|}
 \hline
     \multicolumn{2}{|c|}{\textbf{spectral}} & \multicolumn{3}{c|}{\textbf{COSC}} \\ 
          \multicolumn{2}{|c|}{\emph{Q1}} & \multicolumn{3}{c|}{\emph{Q2}} \\ \cline{1-5}
     STS & COBS & eigen & NumSat & CVCP  \\ 
    \hline
 \textbf{\underline{0.9}} & 0.89 & 0.50 & 0.50 & 0.68 \\
	  0.21 & \textbf{\underline{0.88}} & 0.38 & 0.38 & 0.50 \\	 
	  0.56 & \underline{0.81} & \textbf{0.84} & 0.43 & 0.60 \\
	 	  \textbf{\underline{0.24}} & 0.23 & 0.22 & 0.22 & \textbf{0.24} \\
	 	  \underline{0.81} & 0.79 & \textbf{0.83} & \textbf{0.83} & \textbf{0.83} \\
	 0.04 & \underline{0.65} & \textbf{0.67} & 0.44 & 0.61 \\
	  0.38 & \textbf{\underline{0.94}} & 0.54  & 0.54 & 0.77 \\
 	  0.24 & \textbf{\underline{0.49}}  & 0.15  & 0.15 & 0.26 \\
	  -0.1 & \underline{0.03} & \textbf{0.13} & \textbf{0.13} & 0.11 \\
	  \textbf{\underline{0.17}} & \textbf{\underline{0.17}} & 0.12 & 0.12 & \textbf{0.17} \\
		\hline
	\end{tabular} 
	}

  \end{center}
  \end{table*}

\subsubsection*{Results and conclusion}
The columns of Table \ref{table:hyperparam_selection} marked with \emph{Q1} compare the ARIs obtained with the unsupervised approaches to those obtained with COBS. The best of these two is underlined for each algorithm and dataset combination. Most of the times the constraint-based selection strategy performs better, and often by a large margin. Note for example the large difference for ionosphere: DBSCAN is able to produce a good clustering, but it is only selected using the constraint-based approach. When the unsupervised selection method performs better, the difference is usually small. \emph{We conclude that often the internal measures do not match the actually desired clusters. Constraints provide useful information that can help select a good parameter setting.}

\subsection*{Q2: COBS vs. semi-supervised algorithms}
\label{sec:compare_to_individual_ss}

It is not too surprising that COBS outperforms unsupervised hyperparameter selection, since it has access to more information. We now compare to semi-supervised algorithms, which have access to the same information. \\

\subsubsection*{Existing semi-supervised algorithms}
We compare to the following algorithms, as they are semi-supervised variants of the unsupervised algorithms used in our experiments:
\begin{itemize}
  \setlength\itemsep{-1em}
\item \textbf{MPCKMeans} \cite{Bilenko2004} is a hybrid semi-supervised extension of K-means. It minimizes an objective that combines the within-cluster sum of squares with the cost of violating constraints. This objective is greedily minimized using a procedure based on K-means. Besides a modified cluster assignment step and the usual cluster center re-estimation step, this procedure also adapts an individual metric associated with each cluster in each iteration. We use the implementation available in the WekaUT package\footnote{\url{http://www.cs.utexas.edu/users/ml/risc/code/}}. \\
\item \textbf{FOSC-OpticsDend} \cite{Campello} is a semi-supervised extension of OPTICS, which is in turn based on ideas similar to DBSCAN. The first step of this algorithm is to run the unsupervised OPTICS algorithm, and to construct a dendrogram using its output. The FOSC framework is then used to extract a flat clustering from this dendrogram that is optimal w.r.t. the given constraints.\\
\item \textbf{COSC} \cite{Rangapuram2012} is based on spectral clustering, but optimizes for an objective that combines the normalized cut with a penalty for constraint violation. We use the implementation available on the authors' web page\footnote{\url{http://www.ml.uni-saarland.de/code/cosc/cosc.htm}}. 
\end{itemize}

In our experiments, the only kind of supervision that is given to the algorithms is in the form of pairwise constraints. In particular, the number of clusters $K$ is assumed to be unknown. In COBS, $K$ is treated as any other hyperparameter. MPCKMeans and COSC, however, require specifying the number of clusters. The following strategies are used to select $K$ based on the constraints:
\begin{itemize}
	\item \textbf{NumSat}: We run the algorithms for multiple $K$, and select the clustering that violates the smallest number of constraints. In case of a tie, we choose the solution with the lowest number of clusters. 
	\item \textbf{CVCP}: Cross-Validation for finding Clustering Parameters \cite{Pourrajabi2014} is a cross-validation procedure for semi-supervised clustering. The set of constraints is divided into $n$ independent folds. To evaluate a parameter setting, the algorithm is repeatedly run on the entire dataset given the constraints in $n-1$ folds, keeping aside the $n$th fold as a test set. The clustering that is produced given the constraints in the $n-1$ folds, is then considered as a classifier that distinguishes between must-link and cannot-link constraints in the $n$th fold. The F-measure is used to evaluate the score of this classifier. The performance of the parameter setting is then estimated as the average F-measure over all test folds. This process is repeated for all parameter settings, and the one resulting in the highest average F-measure is retained. The algorithm is then run with this parameter setting using all constraints to produce the final clustering. We use 5-fold cross-validation.
\end{itemize} 

We also compare to unsupervised hyperparameter selection strategies for the semi-supervised algorithms. In particular, we use the silhouette index for MPCKMeans, and the eigengap heuristic for COSC \cite{vonLuxburg}. The affinity matrix for COSC is constructed using local scaling as in \cite{Rangapuram2012}.\\

\subsubsection*{Results and conclusion}
The columns in Table \ref{table:hyperparam_selection} marked with \emph{Q2} show the ARIs obtained with the semi-supervised algorithms. The best result for each type of algorithm (unsupervised or semi-supervised) is indicated in bold. The table shows that in several cases it is more advantageous to use the constraints to optimize the hyperparameters of the unsupervised algorithm (as COBS does). In other cases, it is better to use the constraints within the algorithm itself, to perform a more informed search (as the semi-supervised variants do). \emph{Within the scope of a single clustering algorithm, neither strategy consistently outperforms the other.} For example, if we use spectral clustering on the dermatology data, it is better to use the constraints for tuning the hyperparameters of unsupervised spectral clustering (also varying $k$ and $\sigma$ for constructing the signature matrix) than within COSC, its semi-supervised variant (which uses local scaling for this). In contrast, if we use density-based clustering on the same data, it is better to use constraints in FOSC-OpticsDend (which does not have an $\epsilon$ parameter, and for which $minPts$ is set to 4, a value commonly used in the literature \cite{Ester96adensity-based,Campello}) than to use them to tune the hyperparameters of DBSCAN (varying both $\epsilon$ and $minPts$).

\subsection*{Q3: COBS with multiple unsupervised algorithms}
\label{sec:COBS}
In the previous two subsections, we showed that constraints can be useful to tune the hyperparameters of individual algorithms. Table \ref{table:hyperparam_selection} also shows, however, that no single algorithm (unsupervised or semi-supervised) performs well on all datasets. This motivates the use of COBS to not only select hyperparameters, but also the clustering algorithm. In this subsection we again use Algorithm \ref{COBS_random_algo}, but set $C$ in step 1 now includes clusterings produced by any of the three unsupervised algorithms. \\

\begin{comment}
Due to long runtimes, results for COSC in combination with CVCP are omitted for the two largest datasets (optdigits389 and segmentation) in the experiments that are described next.
  \end{comment}

\begin{figure}
    \captionsetup{justification=centering}
   \includegraphics[width=0.50\textwidth]{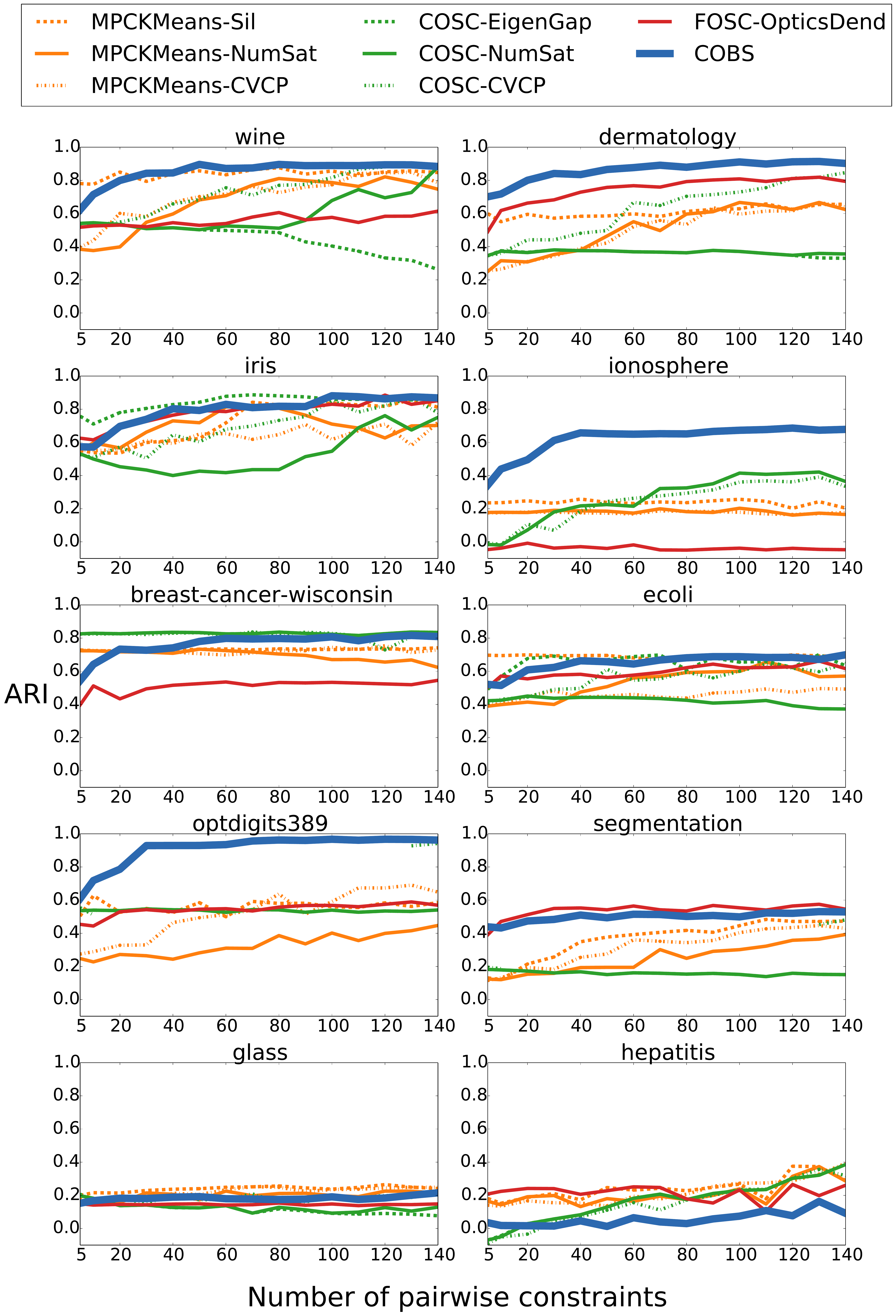}
    \caption{Performance of COBS vs. semi-supervised algorithms}
    \label{fig:COBS_results}
\end{figure}

\subsubsection*{Results}

We compare COBS with existing semi-supervised algorithms in Figure \ref{fig:COBS_results}\footnote{Due to long runtimes of COSC, we do not report results in combination with CVCP on the two largest datasets (optdigits389 and segmentation).}. COBS is able to find relatively good clusterings for the first 8 datasets. While some other approaches also do well on some of these datasets, none of them do so consistently. Compared to each competitor individually, COBS is clearly superior. For example, COSC-EigenGap outperforms COBS on the iris dataset, but performs much worse on several others. COBS performs poorly on glass and hepatitis, as do the other semi-supervised algorithms, although for hepatitis other approaches are able to find better solutions after a larger number of constraints. The overall poor performance on these last two datasets suggests that the class labels do not indicate a natural grouping.\\

Table \ref{table:COBS_vs_best} allows us to assess the quality of the clusterings that are selected by COBS, relative to the quality of the best clustering in the set of generated clusterings. Column 2 shows the highest ARI of all generated clusterings for each dataset. Note that we can only compute this value in an experimental setting, in which we have labels for all elements. In a real clustering application, we cannot simply select the result with the highest ARI. Column 3, then, shows the ARI of the clustering that is actually selected using COBS when it is given 50 constraints. It shows that there still is room for improvement, i.e.\ a more advanced strategy might get closer to the maxima. Nevertheless, even our simple strategy gets close enough to outperform most other semi-supervised methods. The last column of Table \ref{table:COBS_vs_best} shows how often COBS chose a clustering by K-means ('K'), DBSCAN ('D') and spectral clustering ('S'). It illustrates that the selected algorithm strongly depends on the dataset. For example, for ionosphere COBS selects clusterings generated by DBSCAN, as it is the only algorithm able to produce good clusterings of this dataset. For most other datasets, spectral clustering is preferred.\\

\subsubsection*{Conclusion}
If any of the unsupervised algorithms is able to produce good clusterings, COBS can select them using a limited number of constraints. If not, COBS performs poorly, but in our experiments none of the algorithms did well in this case.
\emph{We conclude that it is often better to use constraints to select and tune an unsupervised algorithm, than within a randomly chosen semi-supervised algorithm}.

\begin{comment}
Which algorithm is selected, depends strongly on the dataset. For example, for ionosphere COBS selects clusterings generated by DBSCAN, as that is the only algorithm able to produce relatively good clusterings of this dataset.
\end{comment}

\begin{table}
   \captionsetup{justification=centering}
  \begin{center}

 \caption{ The ARI of the best clustering that is generated by any of the unsupervised algorithms, the ARI of the clustering that is selected after 50 constraints (averaged over 25 runs), and the algorithms that produced the selected clusterings.}\label{table:COBS_vs_best}% Add 'table' caption
 
  \footnotesize
  
  \begin{tabular}{|c|c|c|c|}
 \hline
     \textbf{dataset} & \shortstack{\textbf{best} \\ \textbf{unsupervised}} & \textbf{COBS}  & \shortstack{\textbf{algorithm} \\ \textbf{used}} \\
    \hline
    wine  & 0.93 & 0.90 & K:4/D:0/S:21  \\
	dermatology  & 0.94 & 0.87 & K:12/D:0/S:13   \\
	iris  & 0.88 & 0.80 &  K:9/D:0/S:16   \\
		ionosphere  & 0.7 & 0.65 &  K:0/D:25/S:0   \\
		breast-cancer-wisconsin  & 0.84 & 0.77 & K:4/D:1/S:20   \\
	ecoli  & 0.75 & 0.65 & K:6/D:0/S:19   \\
	optdigits389  & 0.97 & 0.96 & K:0/D:0/S:25   \\
	segmentation  & 0.59 & 0.50 & K:8/D:2/S:15  \\
	hepatitis  & 0.27 & 0.01 & K:1/D:18/S:6  \\
	glass  & 0.29 & 0.19 & K:14/D:0/S:11   \\
	\hline
	\end{tabular}
       
	\end{center}  
   \end{table}

\begin{comment}

Table \ref{table:COBS_vs_best} shows how close COBS gets to what we can maximally attain with a selection strategy. 
Column 2 shows the ARIs of the best clusterings from the generated sets. In practice, we cannot simply select these clusterings, as we would need all labels to calculate the ARIs.  Column 3 shows how close we get to this maximum using COBS with 50 constraints. For most datasets, COBS gets reasonably close to the maximum. However, this is not the case for hepatitis, and to a lesser extent glass, for which all clusterings score relatively low.\\
The last column of Table \ref{table:COBS_vs_best} shows how often COBS chose a clustering by K-means ('K'), DBSCAN ('D') and spectral clustering ('S'). Clusterings produced by different algorithms are selected for different datasets. For example, for ionosphere COBS selects clusterings generated by DBSCAN, as that is the only algorithm able to produce relatively good clusterings of this dataset. 

\end{comment}

\subsection*{Q4: Using COBS with semi-supervised algorithms}
\label{sec:COBS_semisup}
In the previous section we have shown that we can use constraints to do algorithm and hyperparameter selection for \emph{unsupervised} algorithms. On the other hand, constraints can also be useful when used within an adapted clustering procedure, as traditional semi-supervised algorithms do. This raises the question: can we combine both approaches? In this section, we use the constraints to select and tune a \emph{semi-supervised} clustering algorithm. In particular, we vary the hyperparameters of the semi-supervised algorithms to generate the set of clusterings from which we select. The varied hyperparameters are the same as those for their unsupervised variants, except for two. First, $\epsilon$ is not varied for FOSC-OpticsDend, as it is not a hyperparameter for that algorithm. Second, in this section we only use $k$-nearest neighbors graphs for (semi-supervised) spectral clustering, as full similarity graphs lead to long execution times for COSC. \\

\subsubsection*{Results and conclusions}
Column 3 of Table \ref{table:COBS_for_semisup} shows that this strategy does not produce better results. This is caused by using the same constraints twice: once within the semi-supervised algorithms, and once to evaluate the algorithms and select the best-performing one. Obviously, algorithms that overfit the given constraints will get selected in this manner.  \\
The problem could be alleviated by using separate constraints inside the algorithm and for evaluation, but this decreases the number of constraints that can effectively be used for either purpose. Column 4 of Table \ref{table:COBS_for_semisup} shows the average ARIs that are obtained if we use half of the constraints within the semi-supervised algorithms, and half to select one of the generated clusterings afterwards. This works better, but still often not as good as COBS with unsupervised algorithms. Results are only improved for segmentation, hepatitis and glass, the datasets with less clear clustering structure (as indicated by the ARIs).\\
\emph{We conclude that using semi-supervised algorithms within COBS can only be beneficial if the semi-supervised algorithms use different constraints from those used for selection. Even then, when a limited number of constraints is available, using all of them for selection is often the best choice.} 

 \begin{table}
    \captionsetup{justification=centering}

  \begin{center}

 \caption{{\footnotesize ARIs obtained with 50 constraints by COBS with unsupervised algorithms (COBS-U) and with semi-supervised algorithms, with and without splitting the constraint set (COBS-SS and COBS-SS-split). Results are averaged over 25 random constraint sets, except for optdigits389 and segmentation, for which results are averaged over 10 runs.}}\label{table:COBS_for_semisup} 
 {\footnotesize  

\begin{tabular}{|c|c|c|c|}
 \hline
     \textbf{dataset} & \textbf{COBS-U} & \textbf{COBS-SS} & \textbf{COBS-SS-split} \\
    \hline
    wine   & \textbf{0.89} & 0.54 & 0.80  \\
	dermatology   & \textbf{0.85} & 0.62 & 0.81 \\
	iris  & \textbf{0.77} & 0.51 & 0.75 \\
	ionosphere  & \textbf{0.64} & 0.19 & 0.31 \\
	breast-cancer-wisconsin & \textbf{0.79} & 0.50 & 0.69 \\
	ecoli   & \textbf{0.67} & 0.51 & 0.63 \\
	optdigits389 & \textbf{0.92} & 0.51 & 0.80 \\
	segmentation & 0.48 & 0.45 & \textbf{0.54} \\
	hepatitis   & 0.07 & 0.09 & \textbf{0.27}  \\
	glass   & 0.18 & 0.18 & \textbf{0.19} \\
		\hline
	\end{tabular} }
\end{center}
\end{table}

\subsection*{Note on computational complexity}
One might expect COBS to be prohibitively expensive, given the large number of clusterings it needs to generate. This is not the case, for multiple reasons.
 
First, the runtimes of individual clustering algorithms vary greatly, and in addition to that, some semi-supervised algorithms are much slower than their unsupervised counterpart.  As a result, constructing many clusterings with unsupervised algorithms is only slightly more expensive than running the slowest semi-supervised algorithm just once.  In our experiments, for the largest dataset we used (segmentation), generating 931 unsupervised clusterings took 560s on a single core, using scikit-learn implementations.  A single run of COSC, the semi-supervised variant of spectral clustering, took 200s (using the Matlab implementation available on the authors' web page).  If COSC is run multiple times, for instance with different numbers of clusters (as is done in COSC-NumSat and COSC-CVCP), its runtime quickly exceeds that of COBS.

Second, the runtime of COBS can be reduced in several ways.  The cluster generation step can easily be parallelized.  For larger datasets, one might consider doing the algorithm and hyperparameter selection on a sample of the data, and afterwards cluster the complete dataset only once with the selected configuration.

Finally, note that the added cost of doing algorithm and parameter selection is no different from its comparable, and commonly accepted, cost in (semi-)supervised learning. The focus is on maximally exploiting the limited amount of supervision, as obtaining labels or constraints is often expensive, whereas computation is cheap.

\section{Active COBS}
\label{sec:COBS_active}

Obtaining constraints can be costly, as they are often specified by human experts. Consequently, several methods have been proposed to actively select the most informative constraints \cite{Basu2004,Mallapragada2008,Xiong2014}. We first briefly discuss some of these methods, and subsequently present a constraint selection strategy for COBS.

\subsection{Related work}
\label{sec:active_related}
Basu \etal . \cite{Basu2004} were the first to propose an active constraint selection method for semi-supervised clustering. Their strategy is based on the construction of neighborhoods, which are points that are known to belong to the same cluster because must-link constraints are defined between them. These neighborhoods are initialized in the \emph{exploration} phase: $K$ (the number of clusters) instances with cannot-link constraints between them are sought, by iteratively querying the relation between the existing neighborhoods and the point farthest from these neighborhoods. In the subsequent \emph{consolidation} phase these neighborhoods are expanded by iteratively querying a random point against the known neighborhoods until a must-link occurs and the right neighborhood is found. Mallapragada \etal . \cite{Mallapragada2008} extend this strategy by selecting the most uncertain points to query in the consolidation phase, instead of random ones. Note that in these approaches all constraints are queried before the actual clustering is performed.  \\  
More recently, Xiong \etal . \cite{Xiong2014} proposed the normalized point-based uncertainty (NPU) framework. Like the approach introduced by Mallapragada \etal . \cite{Mallapragada2008}, NPU incrementally expands neighborhoods and uses an uncertainty-based principle to determine which pairs to query. In the NPU framework, however, data is re-clustered several times, and at each iteration the current clustering is used to determine the next set of pairs to query. NPU can be used with any semi-supervised clustering algorithm, and Xiong \etal . \cite{Xiong2014} use it with MPCKMeans to experimentally demonstrate its superiority to the method of Mallapragada \etal . \cite{Mallapragada2008}.

\subsection{Active constraint selection in COBS}
\label{sec:active_cobs}
Like the approaches in \cite{Mallapragada2008} and \cite{Xiong2014}, our constraint selection strategy for COBS is based on uncertainty sampling. Defining this uncertainty is straightforward within COBS, because of the availability of a set of clusterings: a pair is more uncertain if more clusterings disagree on whether it should be in the same cluster or not. Algorithm \ref{COBS_active_algo} presents a selection strategy based on this idea. We associate with each clustering $c$ a weight $w_c$ that depends on the number of constraints $c$ was right or wrong about. In each iteration we query the pair with the lowest weighted agreement. The agreement of a pair (line 5 of the algorithm) is defined as the absolute value of the difference between the sum of weights of clusterings in which the instances in the pair belong to the same cluster, and the sum of weights of clusterings in which they belong to a different cluster. The weights of clusterings that correctly ``predict'' the relation between pairs are increased by multiplying with an update factor $m$, weights of other clusterings are decreased by dividing by $m$. As the total number of pairwise constraints is quite large ($\binom{n}{2}$ with $n$ the number of instances), we only consider constraints in a small random sample $P$ of all possible constraints.

\renewcommand{\thealgorithm}{2}

\begin{algorithm}
\caption{Active constraint selection for COBS}
\label{COBS_active_algo}
\begin{algorithmic}[1]
 \REQUIRE $D$: a dataset\\ \ \ \ \ \ \  \emph{budget}: the maximum number of constraints to use \\ \ \ \ \ \ \  \emph{m}: weight update factor  \\ \ \ \ \ \ \  \emph{s}: size of sample of constraints to choose from
\ENSURE a clustering of $D$
\STATE Generate $C$ a set of $s$ clusterings by varying the hyperparameters of several unsupervised clustering algorithms
\STATE Let $w_c = \frac{1}{s}$ for all $c \in C$
\STATE Let $P$ be a sample of all possible pairwise constraints
 \WHILE{$u < \text{\emph{budget}}$}
        \STATE {\footnotesize $(i, j) \gets \displaystyle \argmin_{(p,q) \in P} \bigl| \!\sum_{c \in C}\!\amsbb{I}[c[p]=c[q]]w_{c}\!-\!\!\!\sum_{c \in C}\!\amsbb{I}[c[p]\neq c[q]]w_{c} \bigr| $ }
        \STATE Query pair $(i,j)$
        \STATE $\forall c \in C$: multiply $w_c$ with $m$ if $c$ correctly predicted the \\ \ \ \ \ \ \ \ \ \ relation between $i$ and $j$, divide by $m$ if not
		\STATE $u \gets u + 1$
    \ENDWHILE
\RETURN the clustering with the highest weight  
\end{algorithmic}    
\end{algorithm}

\begin{comment}
\renewcommand{\thealgorithm}{1}

\floatname{algorithm}{Procedure}
\begin{algorithm}
\caption{$MostDisagreedOnPair$}
\label{mostdisagreedon}
\begin{algorithmic}[1]
\REQUIRE $C$: set of clusterings, $\mathbf{w}$: corresponding weights, $P$: sample of pairs
\ENSURE most disagreed on pair from $P$
\FORALL{$(x_p,x_q) \in P$}
    \STATE $s=0, d=0$
    \FORALL{$c_k \in C$}
        \IF{$c_k[p] = c_k[q]$} \STATE $s \gets s + w_k$ 
        	\ELSE \STATE $d \gets d + w_k$
        	\ENDIF
	\ENDFOR
	\STATE calculate agreement for $(x_p,x_q)$ as $|s - d|$ 
\ENDFOR
\RETURN the pair with the lowest agreement
\end{algorithmic}    
\end{algorithm}
\end{comment}

\subsection{Experiments}
We first demonstrate the influence of the weight update factor and sample size, and then compare our approach to active constraint selection with NPU \cite{Xiong2014}. \\

\subsubsection*{Effect of weight update factor and sample size}
\begin{figure}
    \captionsetup{justification=centering}
   \includegraphics[width=0.5\textwidth]{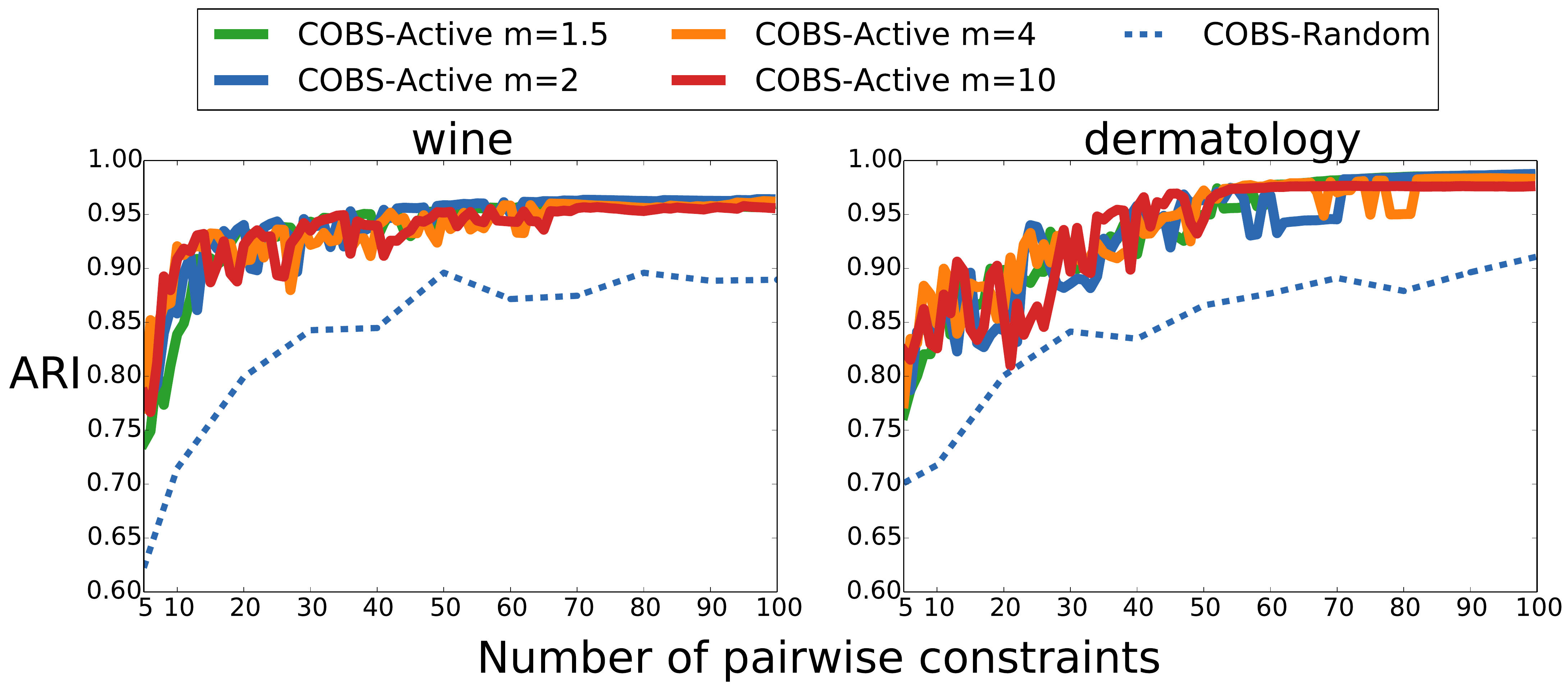}
    \caption{{\footnotesize Active COBS with different weight update factors. The constraint sample size was set to 1000.}}\label{fig:weight_updates}
\end{figure}

\begin{figure}
    \captionsetup{justification=centering}
   \includegraphics[width=0.5\textwidth]{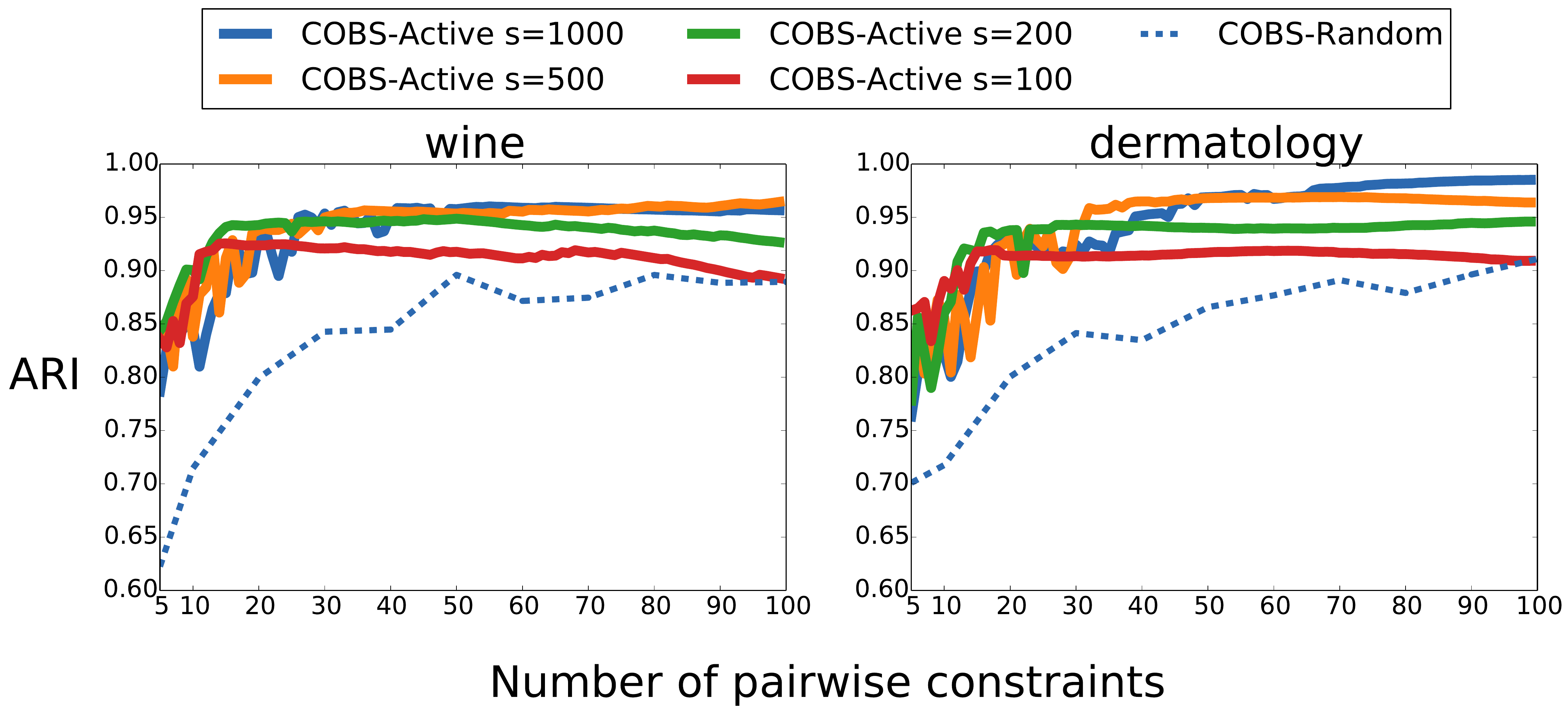}
    \caption{{\footnotesize Active COBS with different sample sizes. The weight update factor was set to 2.}}\label{fig:sample_size}
\end{figure}

Our constraint selection strategy requires specifying a weight update factor $m$ and a sample size $s$. Figure \ref{fig:weight_updates} shows the results for wine and dermatology for various values of $m$. First, the figure shows that the active strategy can significantly improve performance over random selection. Second, it shows that the selection process is not very sensitive to the choice of the update factor. Figure \ref{fig:sample_size} shows the results for various sample sizes. It shows that the sample size has a limited effect on performance for a small number of constraints, but that this effect increases as more constraints are given. In the remainder of this section we use a sample of 1000 constraints (i.e.\ we try to choose the most useful constraints to ask from 1000 possible queries), and set the weight update factor to 2.   \\

\begin{figure}
    \captionsetup{justification=centering}
   \includegraphics[width=0.5\textwidth]{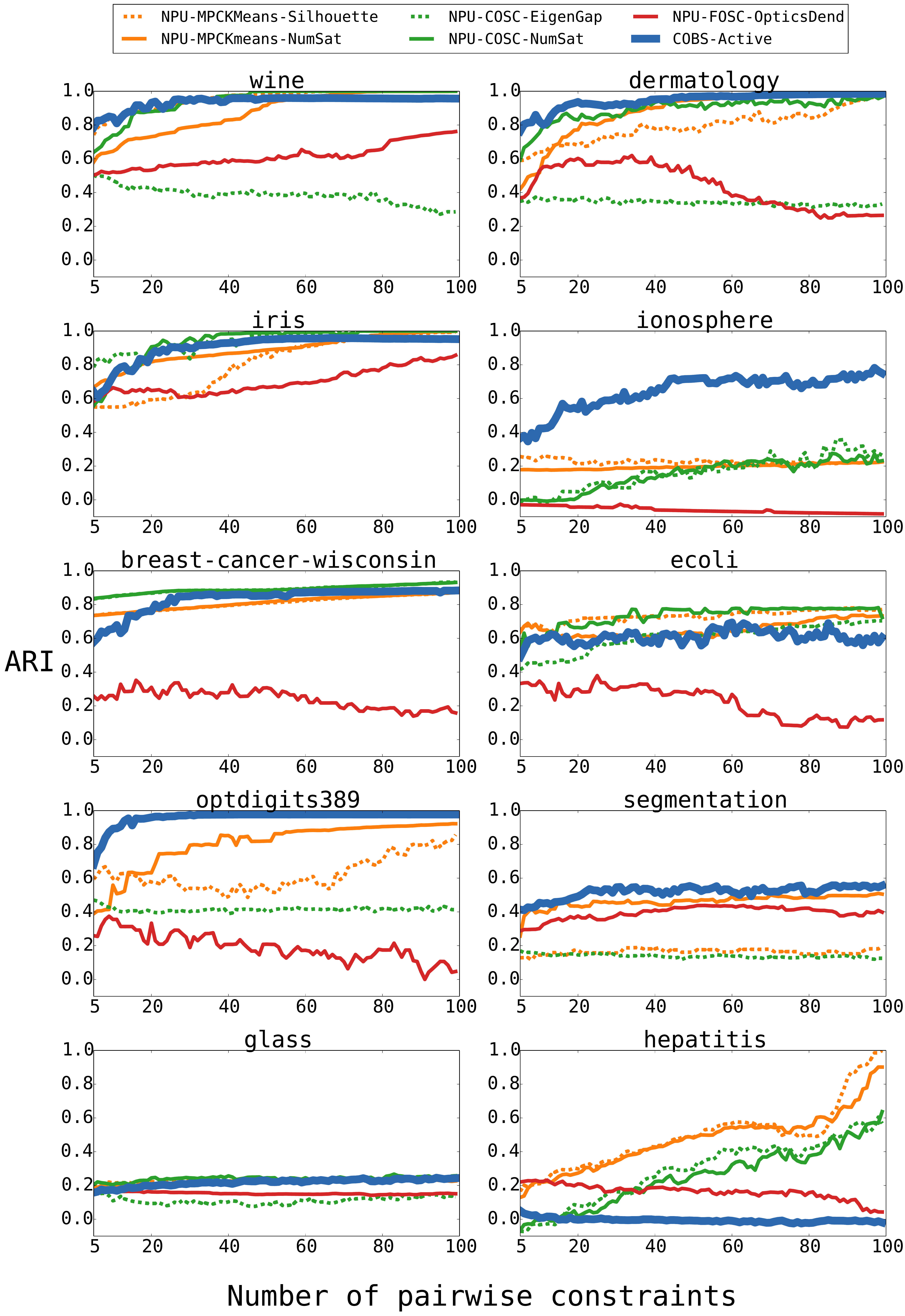}
    \caption{{\footnotesize Comparison of active COBS to NPU in combination with different semi-supervised clustering algorithms }}
    \label{fig:COBS_results_active}
\end{figure}

\subsubsection*{Comparison to active selection with  NPU}
 NPU \cite{Xiong2014} can be used in combination with any semi-supervised clustering algorithm, we use the same ones as in the previous section. We do not include CVCP hyperparameter selection in these experiments, because of its high computational complexity (for these experiments we cannot cluster for several fixed numbers of constraints, as the choice of the next constraints depends on the current clustering). For the same reason we only include the EigenGap parameter selection method for the two largest datasets (opdigits389 and segmentation) in these experiments. The results are shown in Figure \ref{fig:COBS_results_active}. For the first 8 datasets, the conclusions are similar to those for the random setting: COBS consistently performs relatively well. Also in the active setting, none of the approaches produces a clustering with a high ARI for glass. For hepatitis, however, MPCKMeans is able to find good clusterings while COBS is not, albeit only after a relatively large number of constraints (hepatitis contains 112 instances). This implies that, although the labels might not represent a natural grouping, the class structure does match the bias of MPCKMeans, and given many constraints the algorithm finds this structure.\\

\subsubsection*{Time complexity}
We distinguish between the offline and online stages of COBS. In the offline stage, the set of clusterings is generated. As mentioned before, this took 560s on a single core for the largest dataset (segmentation, with 2100 instances). In the online stage, we select the most informative pairs and ask the user about their relation. Execution time is particularly important here, as this stage requires user interaction. In active COBS, selecting the next pair to query is  $\mathcal{O}(|C||P|)$, as we have to loop through all clusterings ($|C|$) for each constraint in the sample ($|P|$). For the setup used in our experiments ($|C|=931$, $|P|=1000$), this was always less than 0.02s. Note that this time does not depend on the size of the dataset (as all clusterings are generated beforehand). In contrast, NPU requires re-clustering the data several times during the constraint selection process, which is usually significantly more computationally expensive. \\

\subsubsection*{Conclusion}
The COBS approach allows for a straightforward definition of uncertainty: pairs of instances are more uncertain if more clusterings disagree on them. Selecting the most uncertain pairs first can significantly increase performance.

\section{Conclusion}
\label{sec:conclusion}

Exploiting constraints has been the subject of substantial research, but all existing methods use them within the clustering process of individual algorithms. In contrast, we propose to use them to choose between clusterings generated by different unsupervised algorithms, ran with different parameter settings. We experimentally show that this strategy is superior to all the semi-supervised algorithms compared to, which themselves are state of the art and representative for a wide range of algorithms. For the majority of the datasets, it works as well as the best among them, and on average it performs much better. The generated clusterings can also be used to select more informative constraints first, which further improves performance.

In future work, we would like to study several strategies that have been used in supervised learning in the context of semi-supervised clustering. In particular, we want to consider more advanced algorithm and hyperparameter optimization strategies (as in \cite{AutoWEKA}), meta-learning approaches (as in \cite{Brazdil}), and combinations of these two (as in \cite{NIPS2015_5872}).

\bibliographystyle{plain}
\bibliography{references}

\end{document}